\documentclass[10pt,twocolumn,letterpaper]{article}

\usepackage{cvpr}              % To produce the CAMERA-READY version
\usepackage{amsmath}
\usepackage{amsfonts}
\usepackage{multirow}
\usepackage[accsupp]{axessibility}

\definecolor{cvprblue}{rgb}{0.21,0.49,0.74}
\usepackage[pagebackref,breaklinks,colorlinks,allcolors=cvprblue]{hyperref}

\def\paperID{7} % *** Enter the Paper ID here
\def\confName{CVPR}
\def\confYear{2026}

\title{Combining Boundary Supervision and Segment-Level Regularization for Fine-Grained Action Segmentation}

\author{Hinako Mitsuoka \qquad Kazuhiro Hotta\\
Meijo University\\
1-501 Shiogamaguchi, Tempaku-ku, Nagoya 468-8502, Japan\\
{\tt\small 263441509@ccmailg.meijo-u.ac.jp, kazuhotta@meijo-u.ac.jp}
}

\begin{document}
\maketitle
\begin{abstract}
Recent progress in Temporal Action Segmentation (TAS) has increasingly relied on complex architectures, which can hinder practical deployment.
We present a lightweight dual-loss training framework that improves fine-grained segmentation quality with only one additional output channel and two auxiliary loss terms, requiring minimal architectural modification.
Our approach combines a boundary-regression loss that promotes accurate temporal localization via a single-channel boundary prediction and a CDF-based segment-level regularization loss that encourages coherent within-segment structure by matching cumulative distributions over predicted and ground-truth segments.
The framework is architecture-agnostic and can be integrated into existing TAS models (e.g., MS-TCN, C2F-TCN, FACT) as a training-time loss function.
Across three benchmark datasets, the proposed method improves segment-level consistency and boundary quality, yielding higher F1 and Edit scores across three different models.
Frame-wise accuracy remains largely unchanged, highlighting that precise segmentation can be achieved through simple loss design rather than heavier architectures or inference-time refinements.
\end{abstract}

\section{Introduction}
\label{sec:introduction}

\begin{figure}[t]
\centering
\includegraphics[width=1.0\columnwidth]{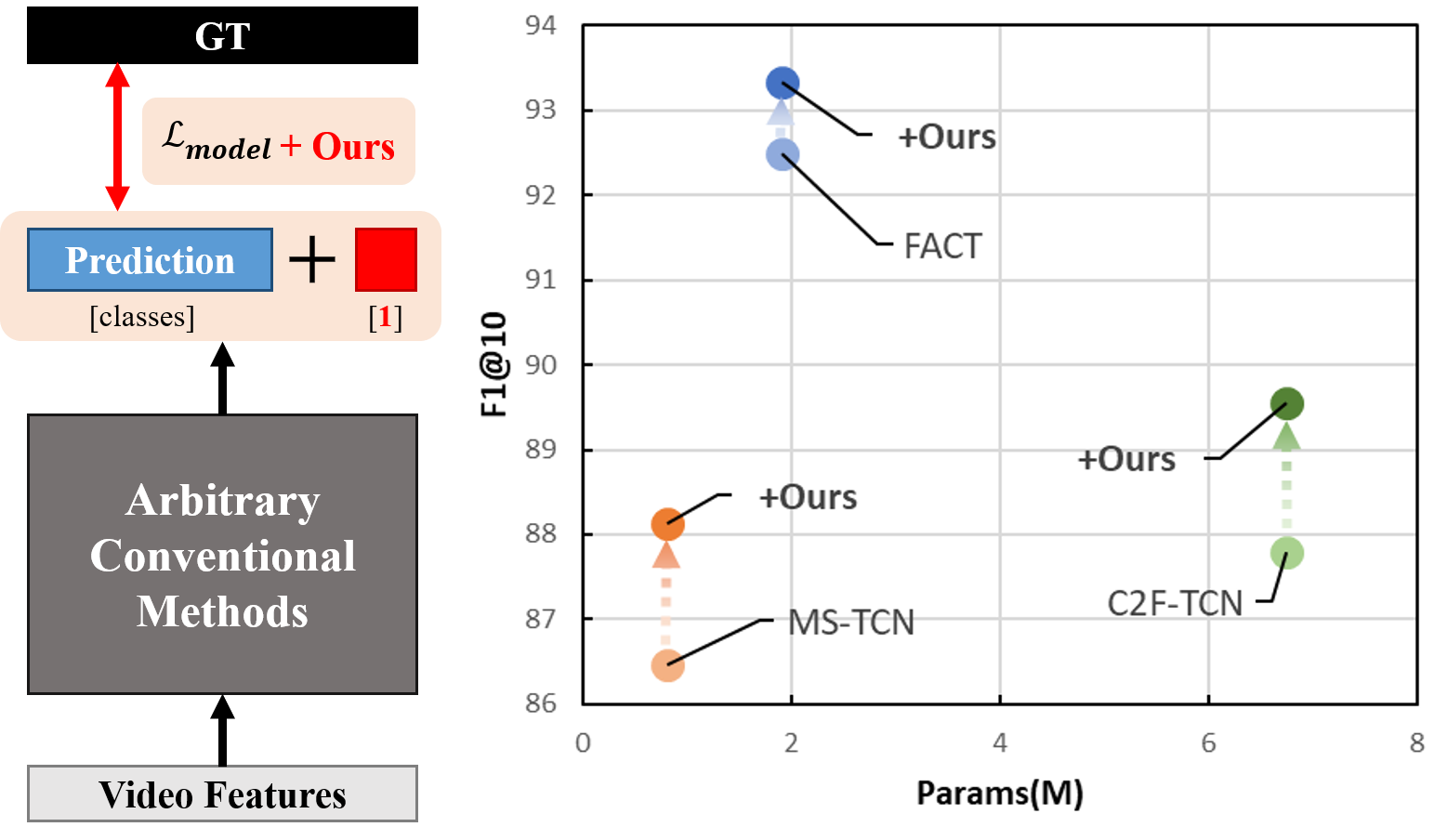} 
\caption{
\textbf{Overview of the proposed method and its performance improvement.}
The left shows that the proposed method adds a single boundary output channel and introduces two auxiliary training losses that can be integrated into conventional TAS models with minimal modification.
The right shows the relationship between the F1 score and the number of parameters on GTEA, illustrating improved performance with minimal parameter overhead.
}
\label{fig1}
\end{figure}

Recent advances in Temporal Action Segmentation (TAS) have been driven by increasingly complex architectures \cite{mstcn, asformer, ltc2023bahrami, liu2023diffusion, uvast, fact}. 
The field has evolved from early single-path models such as MS-TCN \cite{mstcn} to multi-path frameworks like ASRF \cite{asrf} and FACT \cite{fact}, which employ auxiliary modules for boundary refinement and other tasks. 
Although effective, such designs increase computational complexity and can reduce interpretability. 
While frame-wise classification losses such as cross-entropy remain a core component in many supervised TAS pipelines, they often lead to over-segmentation and boundary misalignment due to unstable frame-level predictions and ambiguous label transitions.

We propose a simple yet effective training-time enhancement for fully supervised TAS by augmenting existing models with two auxiliary losses and only one additional boundary output channel, as shown in Figure \ref{fig1} (left).
The first loss is a boundary-regression objective that enables accurate temporal boundary localization via a single class-agnostic boundary prediction, without introducing auxiliary branches.
The second is a CDF-based segment shape regularization loss, which acts as a lightweight segment-level structural prior by penalizing discrepancies between cumulative distributions computed from predictions and ground-truth segments, thereby mitigating over-segmentation and boundary errors.
Unlike post-hoc refinement approaches, our method injects structural guidance directly into supervised training, requiring no inference-time refinement.
Overall, the proposed losses are architecture-agnostic and can be integrated into existing TAS backbones with minimal implementation overhead.

We evaluate the proposed method on three widely used benchmarks: GTEA \cite{gtea}, 50Salads \cite{50salads}, and Breakfast \cite{breakfast}, with three representative architectures: MS-TCN, C2F-TCN \cite{c2ftcn}, and FACT. 
The proposed method achieved consistent improvements across all datasets and architectures, confirming its general effectiveness.
Figure \ref{fig1} (right) presents a representative example.
Furthermore, the proposed method achieved a better trade-off between accuracy and complexity than boundary-regression methods such as ASRF \cite{asrf} and BCN \cite{bcn}, and is complementary to post-hoc refinements such as ASOT \cite{asot}.

In summary, our main contributions are as follows.
\begin{itemize}
\item We propose a lightweight dual-loss training framework for fully supervised TAS that combines boundary supervision with a segment-level regularization loss.
\item Our method is \textbf{architecture-agnostic}, requiring only one additional boundary output channel and two auxiliary loss terms as a training-time objective.
\item We introduce a CDF-based segment shape regularization loss and a decoupled loss assignment that applies boundary and segment regularization to different temporal regions to reduce optimization conflicts.
\item When our method is used with MS-TCN, our method improved the performance by up to 4.6\% in Edit and 5.4\% in F1@10 on benchmark datasets, while requiring only minimal architectural modification.
\end{itemize}

\section{Related Works}
\label{sec:related}

\subsection{Temporal Action Segmentation}
\label{subsub:architecture}

Temporal Action Segmentation (TAS) has evolved from hand-crafted pipelines to deep learning architectures. 
Early approaches employed sliding windows \cite{6247801, karaman2014fast}, probabilistic models such as HMMs \cite{kuehne2016end, richard2017weakly}, and RNNs \cite{singh2016multi}, but struggled to capture long-range dependencies and maintain temporal resolution. 
Inspired by the success of WaveNet in speech modeling \cite{wavenet}, Temporal Convolutional Networks (TCNs) became a widely adopted backbone for TAS, leading to methods such as ED-TCN \cite{edtcn}, MS-TCN \cite{mstcn}, and MS-TCN++ \cite{mstcn++}, C2F-TCN \cite{c2ftcn}. 
More recent work leverages Transformers \cite{vit} and multi-scale fusion, e.g., FACT \cite{fact}, to further boost accuracy. 
However, these improvements often increase computational overhead, parameter counts, and latency, limiting deployment in real-world scenarios.

While frame-wise classification losses (e.g., cross-entropy) remain a core component in many supervised TAS pipelines, recent approaches often incorporate additional objectives or refinement mechanisms to improve temporal coherence.
Nevertheless, over-segmentation and boundary misalignment can persist when explicit segment-level structural guidance is limited.
Our work focuses on loss-level enhancements that introduce boundary-aware and segment-level supervision with minimal architectural modification.

\subsection{Boundary-Regression Methods}
\label{subsub:boundary}

Several works have tackled over-segmentation and imprecise transitions by explicitly modeling boundaries. 
ASRF \cite{asrf} employs a two-branch design, combining an Action Segmentation Branch with a Boundary Regression Branch that predicts class-agnostic boundaries and refines predictions via post-processing. 
BCN \cite{bcn} adopts a cascaded framework where each stage refines frame-wise predictions using both the input features and outputs of previous stages, and incorporates a Barrier Generation Module that predicts boundary probabilities and applies local barrier pooling as a temporal regularizer to frame-level features, improving transition alignment.
Although these approaches are effective, they require dedicated branches or heads, increasing architectural complexity and inference cost.

In contrast, our method introduces boundary supervision in a lightweight manner by adding a single class-agnostic boundary output channel and a corresponding training loss.
This design enables boundary-sensitive representations without auxiliary branches, thereby integrating seamlessly into existing TAS models with minimal overhead.

\begin{figure*}[t]
\centering
\includegraphics[width=0.6\textwidth]{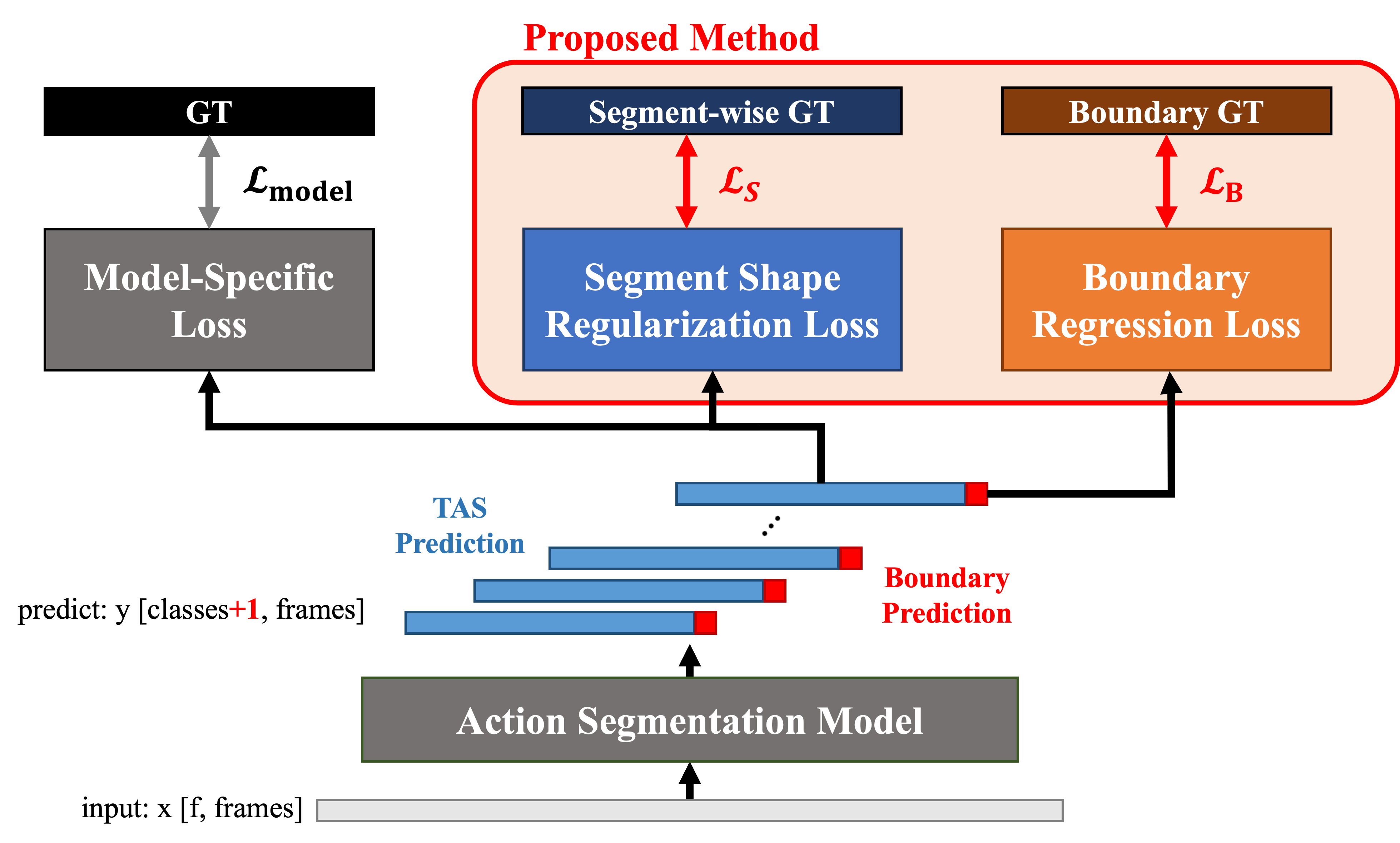}
\caption{\textbf{Overview of the proposed training framework.}
In addition to the model-specific loss, our method introduces two complementary auxiliary losses: a boundary-regression loss using an additional boundary output channel and a CDF-based segment shape regularization loss.
The losses are applied selectively to different temporal regions (boundary vs non-boundary) to reduce optimization conflicts, and can be combined with existing TAS models with minimal architectural modification.
}
\label{fig2}
\end{figure*}

\subsection{Segment-Level Regularization for Temporal Alignment}
\label{subsub:ot}

Segment-level constraints have been explored to improve temporal coherence in TAS, including post-hoc refinement strategies and objectives that encourage structured predictions.
For example, ASOT \cite{asot} and VASOT \cite{vasot} leverage optimal transport to align frame-level representations with latent prototypes in unsupervised or self-supervised settings, and ASOT can also refine supervised predictions as a post-hoc step.
Such inference-time refinements, however, operate outside the supervised training objective and therefore do not directly shape the model’s representations during learning.

In contrast, our method introduces a CDF-based segment shape regularization loss as a training-time objective.
By penalizing discrepancies between cumulative distributions computed from predictions and ground-truth segments, the proposed loss encourages coherent within-segment structure and mitigates over-segmentation.
Although this formulation is inspired by the connection between 1D Wasserstein distance and cumulative distributions, we use it as a simple supervised segment-level shape constraint rather than an explicit OT matching procedure.
Importantly, this regularization is complementary to post-hoc refinement methods: it improves temporal coherence during training, while refinement can optionally further adjust predictions at inference.

\begin{figure}[t]
\centering
\includegraphics[width=0.85\columnwidth]{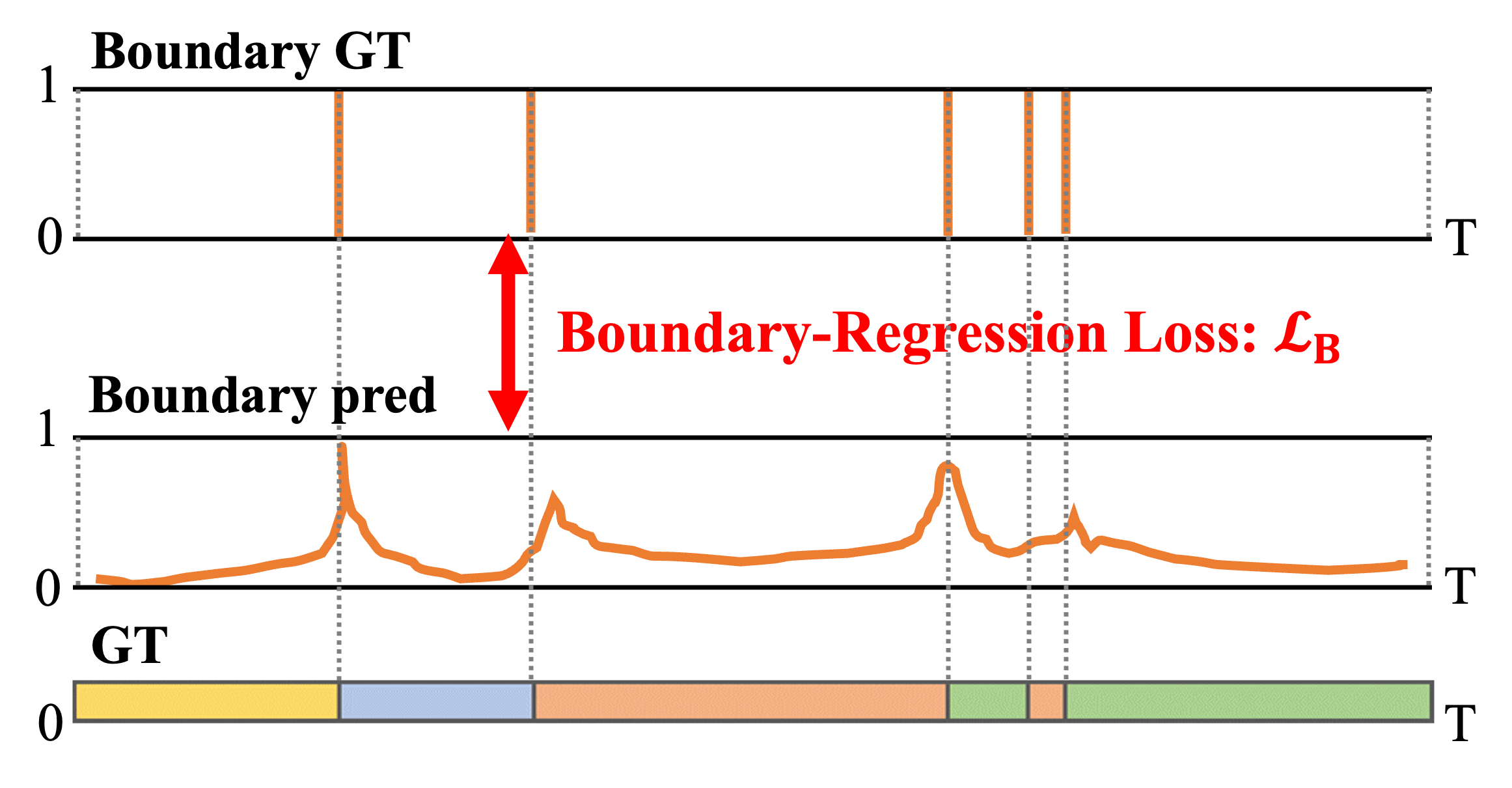} 
\caption{Visualization of \textbf{Boundary-Regression Loss $\mathcal{L}_B$}.  
The additional output channel predicts a boundary probability curve (\textcolor{orange}{orange}), which is supervised by a binary ground-truth boundary mask (top).
The loss encourages high responses around class transitions while maintaining low values elsewhere.  
Segment labels (bottom) are color-coded for clarity.
}
\label{fig3}
\end{figure}
\section{Methodology}
\label{sec:methodology}

We propose a simple training-time enhancement for fully supervised TAS that can be integrated into existing architectures with minimal modification.
The overview is shown in Figure \ref{fig2}.
Our method introduces two complementary auxiliary losses, targeting common issues in TAS: over-segmentation from unstable frame-wise predictions, and degraded segment-level structure.
We describe each component and its joint effect below.

\subsection{Boundary-Regression Loss}
\label{sub:boundary}

To improve localization of action transitions, we add a \textbf{single class-agnostic boundary channel}, as illustrated in Figure \ref{fig2}.  
Conventional TAS models output a tensor of size $C \times T$, where $C$ is the number of classes and $T$ is the number of frames.  
We extend this to $(C{+}1) \times T$ by introducing an additional boundary output.

This additional channel outputs a boundary probability $\hat{b}_t \in [0,1]$ for each frame, supervised by a binary boundary mask $b_t$ derived from ground-truth transitions, as illustrated in Figure \ref{fig3}.
The boundary supervision is formulated as a binary cross-entropy loss
\begin{equation}
\mathcal{L}_{\text{B}} = - \frac{1}{T} \sum_{t=1}^{T} \big( b_t \log \hat{b}_t + (1 - b_t) \log (1 - \hat{b}_t) \big).
\end{equation}

This requires only a minimal modification to the output dimension, without auxiliary branches or notable overhead.

\subsection{Segment Shape Regularization Loss}
\label{sub:emd}

\begin{figure}[t]
\centering
\includegraphics[width=0.85\columnwidth]{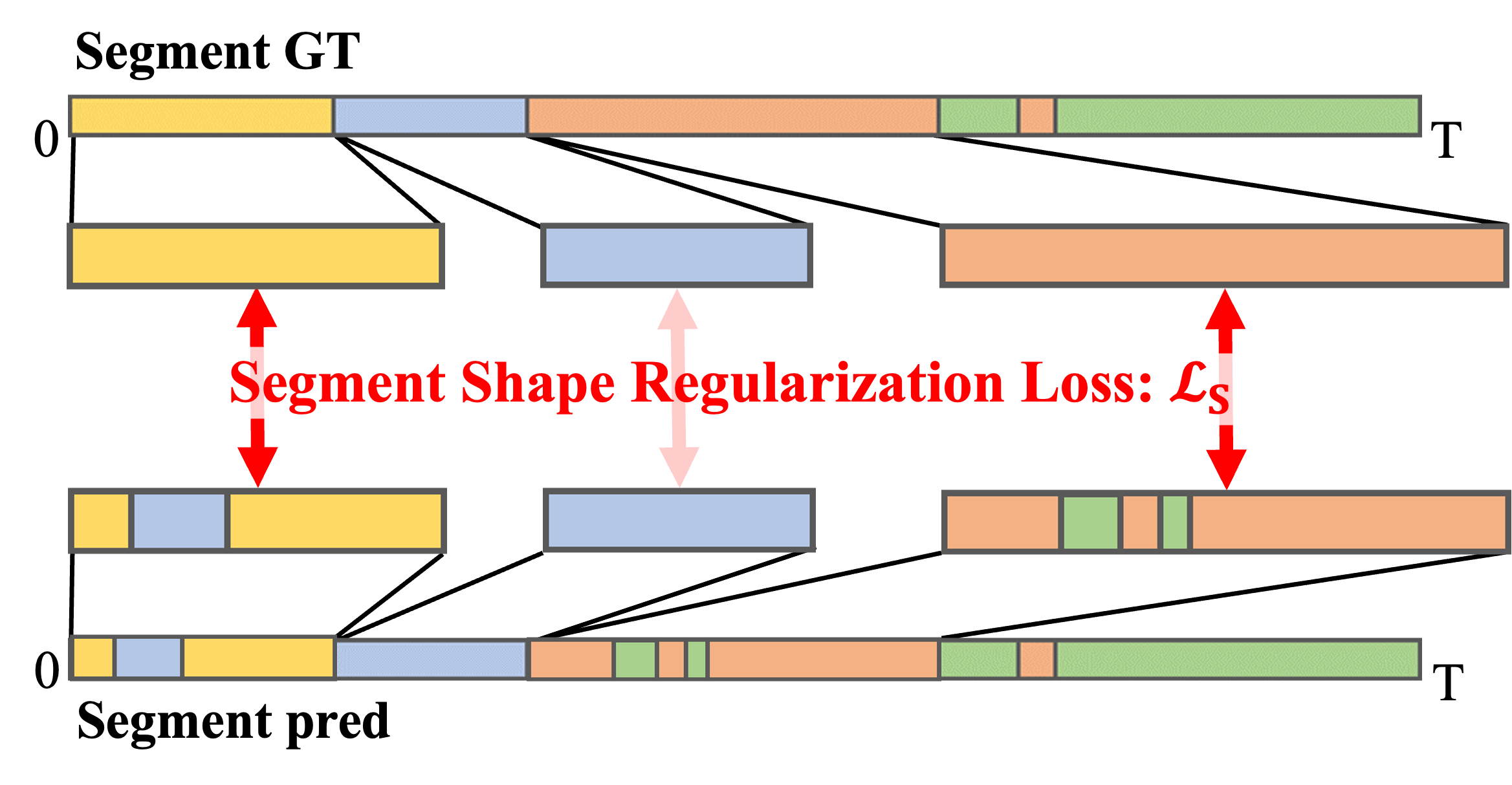} 
\caption{Visualization of the \textbf{CDF-based segment shape regularization loss $\mathcal{L}_{\text{S}}$}.  
Each ground-truth segment (top) is compared with its corresponding predicted region (bottom) using cumulative distributions.
The loss penalizes structural mismatches such as over-segmentation or fragmented predictions by measuring discrepancies between cumulative probability distributions within each segment.
}
\label{fig4}
\end{figure}

Although the boundary loss encourages sharper transitions at action boundaries, it does not explicitly enforce structural consistency within segments.
To complement this, we introduce a \textbf{CDF-based segment shape regularization loss}, which encourages coherent within-segment structure by aligning cumulative distributions computed from predictions and ground-truth segments, as illustrated in Figure \ref{fig4}.
This loss acts as a lightweight segment-level structural prior and is activated after coarse frame-level predictions have stabilized, making it effective at mitigating residual errors such as over-segmentation and local misalignments.

Let $\hat{Y} \in \mathbb{R}^{C \times T}$ denote the predicted class probability map obtained via softmax.  
For each ground-truth segment $S_i = [s_i:e_i]$ with class label $c_i$, we exclude a $\delta$-frame margin near boundaries to avoid unstable supervision around ambiguous class-transition frames where labels change abruptly, and obtain a subsequence of length $L_i = e_i - s_i - 2\delta$.  
If $L_i \leq 0$, the segment is skipped.  
We define the predicted distribution as
\begin{equation}
\hat{p}_i = \mathrm{norm}_{\ell_1}\big(\hat{Y}[c_i,\, s_i{+}\delta:e_i{-}\delta]\big) \in \mathbb{R}^{L_i}.
\end{equation}
and the ground-truth distribution as a uniform vector $p_i = \tfrac{1}{L_i}\mathbf{1}_{L_i} \in \mathbb{R}^{L_i}$.

The loss is computed as the squared $\ell_2$ distance between their cumulative distributions.
\begin{equation}
\mathcal{L}_{\text{S}} = \frac{1}{N} \sum_{i=1}^{N} \frac{1}{L_i}
\sum_{j=1}^{L_i} \left( \text{CDF}(\hat{p}_i)[j] - \text{CDF}(p_i)[j] \right)^2.
\end{equation}
where $\text{CDF}(\cdot)$ denotes the cumulative distribution function, which accumulates probabilities along the temporal axis.  
This formulation is inspired by the connection between the 1D Wasserstein distance and cumulative distributions, and serves as a lightweight segment-level shape constraint. 
By aligning cumulative distributions, this formulation encourages predicted probabilities to match the temporal shape of ground-truth segments while remaining computationally efficient.  
In practice, $\mathcal{L}_{\text{S}}$ is activated after a warm-up period (e.g., $E_{\text{start}}{=}20$ epochs for MS-TCN).

\subsection{Loss Assignment to Avoid Gradient Conflicts}
\label{sub:hybrid}

Directly applying both $\mathcal{L}_{\text{B}}$ and $\mathcal{L}_{\text{S}}$ over the entire sequence can introduce conflicting gradients:
$\mathcal{L}_{\text{B}}$ favors sharp changes around transitions, while $\mathcal{L}_{\text{S}}$ encourages coherent within-segment structure, often resulting in unstable optimization.

To avoid this, we adopt a \textbf{temporally decoupled loss assignment}. 
The temporal axis is partitioned into \textbf{boundary regions} (a small window, e.g., $\pm5$ frames around ground-truth transitions) and \textbf{non-boundary regions} (all other frames). 
$\mathcal{L}_\text{B}$ is applied only within boundary regions to sharpen transitions, while $\mathcal{L}_\text{S}$ is applied to non-boundary regions to improve intra-segment consistency.

The combined loss is thus formulated as
\begin{equation}
\mathcal{L}_{\text{proposed}} = \lambda_{\text{B}} \mathcal{L}_{\text{B}}^{\text{(boundary)}} + \lambda_{\text{S}} \mathcal{L}_{\text{S}}^{\text{(non-boundary)}}
\end{equation}
where $\lambda_{\text{B}}$ and $\lambda_{\text{S}}$ balance the two components.

Finally, the proposed auxiliary losses are added to the standard model-specific loss $\mathcal{L}_{\text{model}}$ (e.g., frame-wise cross-entropy), yielding the overall objective as
\begin{equation}
\mathcal{L}_{\text{total}} = \mathcal{L}_{\text{model}} + \mathcal{L}_{\text{proposed}}.
\end{equation}
This formulation ensures that boundary localization and segment-level shape consistency are optimized in complementary regions of the sequence.
\section{Experiments}
\label{sec:experiments}

We evaluate the effectiveness of the proposed method on three benchmark datasets: Georgia Tech Egocentric Activities (GTEA) \cite{gtea}, 50Salads \cite{50salads}, and Breakfast \cite{breakfast}.
To verify its general applicability, we apply our method to three structurally different models: MS-TCN \cite{mstcn}, C2F-TCN \cite{c2ftcn}, and FACT \cite{fact}.
We also compare the proposed method with conventional boundary-aware approaches \cite{asrf, bcn} to highlight its efficiency--accuracy trade-off.

\subsection{Experimental Conditions}
\label{sub:conditions}

\subsubsection{Datasets.}
\label{subsub:datasets}

We evaluate the proposed method on three benchmark datasets. 
\textbf{GTEA} comprises 28 egocentric videos (7 activities, 11 classes) recorded from 4 subjects, and we use a leave-one-subject-out cross-validation protocol.  
\textbf{50Salads} contains 50 long videos (17 classes) of salad preparation, evaluated by 5-fold cross-validation.  
\textbf{Breakfast} is a large-scale benchmark with 1,716 long videos from 10 recipes and 48 action classes, assessed by the standard 4-fold cross-validation protocol.
Thus, all experiments are conducted using cross-validation protocols recommended in prior work, and the reported scores are averaged across all folds.

For all datasets, we utilize I3D \cite{i3d} features extracted from video frames as input.  
For GTEA and Breakfast, we retain the original temporal resolution of 15 frames per second (fps).  
For 50Salads, the features are downsampled from 30 fps to 15 fps for consistency across datasets.  
This process follows the conventional settings used in prior works \cite{mstcn, asrf}.

\subsubsection{Metrics.}
\label{subsub:metrics}

We evaluate model performance by three standard metrics in TAS: frame-wise accuracy (\textit{Acc}), segmental edit distance (\textit{Edit}), and the segmental F1 score (\textit{F1}). 
The F1 score is computed at intersection-over-union (IoU) thresholds of 10\%, 25\%, and 50\%, denoted as F1@\{10, 25, 50\}.  

Frame-wise accuracy measures the proportion of correctly classified frames, but it is biased toward longer segments and underestimates over-segmentation errors. 
Therefore, following prior works \cite{mstcn, asrf, fact}, we emphasize segmental F1, which better captures segmental quality and temporal consistency. 
Edit distance further evaluates temporal alignment by measuring the number of operations required to match predicted and ground-truth sequences.

\begin{table}[t]
\centering
\resizebox{0.9\columnwidth}{!}{
\begin{tabular}{l|ccccc}
\hline
Methods & Dataset & $\lambda_{\text{B}}$ & $\lambda_{\text{S}}$ & $E_{\text{start}}$ & $E_{\text{max}}$ \\
\hline
\hline
\multirow{3}{*}{MS-TCN} & GTEA & $1.0 \times 10^{-4}$  & \multirow{3}{*}{$1.0 \times 10^{-3}$} & \multirow{3}{*}{20} & \multirow{3}{*}{50}  \\
& 50Salads & $1.0 \times 10^{-3}$  &  &  &   \\
& Breakfast & $1.0 \times 10^{-4}$  &  &  &   \\
\hline
\multirow{3}{*}{C2F-TCN} & GTEA & $1.0 \times 10^{-4}$  & \multirow{3}{*}{$1.0 \times 10^{-3}$} & \multirow{3}{*}{200} & \multirow{3}{*}{500}  \\
& 50Salads & $1.0 \times 10^{-3}$  &  &  &   \\
& Breakfast & $1.0 \times 10^{-5}$  &  &  &   \\
\hline
\multirow{3}{*}{FACT} & GTEA & $1.0 \times 10^{-4}$ & \multirow{3}{*}{$1.0 \times 10^{-4}$} & 100 & 400 \\
 & 50Salads & $5.0 \times 10^{-3}$ &  & 100 & 400 \\
 & Breakfast & $5.0 \times 10^{-3}$ &  & 40 & 150 \\
\hline
\end{tabular}
}
\caption{Hyperparameters used for each model and dataset.}
\label{table1}
\end{table}

\subsubsection{Implementation Details}
\label{subsub:imp}

We implemented our models in PyTorch and trained them on an NVIDIA RTX A6000 Ada GPU. 
All baseline models were implemented and evaluated by following the settings and protocols described in their original papers.

For fair comparison, we adopted the official training settings of each backbone model (MS-TCN, C2F-TCN, and FACT), 
including optimizer configuration, learning schedule, and data augmentation. 
The maximum number of training epochs ($E_{\text{max}}$) was set by following prior works: 50 for MS-TCN, 500 for C2F-TCN, and 400/150 for FACT on GTEA and 50Salads / Breakfast, respectively.

The proposed Boundary-Regression Loss ($\mathcal{L}_{\text{B}}$) and CDF-based Segment Shape Regularization Loss ($\mathcal{L}_{\text{S}}$) were added to each model's original training objective.
We used a boundary window size of $\pm5$ frames and set the segment margin $\delta=5$ in all experiments.
The loss weights and warm-up epochs were selected by cross-validation for each model and dataset, as summarized in Table~\ref{table1}.
No additional model-specific tuning was performed beyond the loss configuration.

\begin{table*}[t]
\centering
% \scalebox{0.85}{
\resizebox{2.\columnwidth}{!}{
\begin{tabular}{lccccc |ccccc |ccccc}
\hline
Dataset & \multicolumn{5}{c|}{GTEA} & \multicolumn{5}{c|}{50Salads} & \multicolumn{5}{c}{Breakfast} \\
\hline
Methods & \multicolumn{3}{c}{F1@\{10,25,50\}} & Edit & Acc & \multicolumn{3}{c}{F1@\{10,25,50\}} & Edit & Acc & \multicolumn{3}{c}{F1@\{10,25,50\}} & Edit & Acc \\
\hline
\hline
\textit{For Standard Architecture} \\
\hline
MS-TCN \cite{mstcn} 
& 86.47 & 83.73 & 71.20 & 79.92 & 76.53
& 73.96 & 71.42 & 62.12 & 66.29 & 79.11
& 50.61 & 46.41 & 36.58 & 61.75 & 66.96 \\
+ Ours 
& \textbf{88.13} & \textbf{84.62}  & \textbf{72.40} & \textbf{83.23} & \textbf{78.20} 
& \textbf{77.44} & \textbf{75.22} & \textbf{67.31} & \textbf{70.29} & \textbf{81.17}
& \textbf{55.96} & \textbf{51.39} & \textbf{40.78} & \textbf{62.29} & \textbf{67.46} \\
\hline
\textit{For Complex Architecture} \\
\hline
C2F-TCN \cite{c2ftcn} & 87.80 & 86.38 & 76.72 & 81.14 & \textbf{80.32} 
& 84.03 & \textbf{82.65} & 74.37 & 76.76 & \textbf{86.63}
& 71.41 & 67.80 & 57.28 & 67.84 & \textbf{76.10} \\
+ Ours & \textbf{89.55} & \textbf{88.64} & \textbf{78.75} & \textbf{85.75} & 80.26 
& \textbf{85.20} & 81.97 & \textbf{74.97} & \textbf{78.55} & 86.18
& \textbf{72.40} & \textbf{68.59} & \textbf{57.69} & \textbf{68.54} & 76.03 \\
\hline
FACT \cite{fact} & 92.49 & 90.36 & 79.83 & 89.89 & 84.74 
& 82.89 & 80.94 & 75.52 & 76.70 & 85.47
& 80.36 & 75.79 & 65.47 & 79.23 & 75.89 \\
+ Ours & \textbf{93.34} & \textbf{91.77} & \textbf{81.40} & \textbf{91.34} & \textbf{85.33} 
& \textbf{83.17} & \textbf{81.29} & \textbf{76.56} & \textbf{77.56} & \textbf{86.44}
& \textbf{81.36} & \textbf{76.49} & \textbf{65.98} & \textbf{79.44} & \textbf{76.08} \\
\hline
\end{tabular}
}
\caption{\textbf{Performance comparison on GTEA, 50Salads and Breakfast.}
We apply our method to MS-TCN, C2F-TCN, and FACT.}
\label{table2}
\end{table*}

\begin{table*}[t]
\centering
\scalebox{0.85}{
% \resizebox{.95\columnwidth}{!}{
\begin{tabular}{lccccccc}
\hline
Methods & \multicolumn{3}{c}{F1@\{10,25,50\}} & Edit & Acc & MACs(G) & Params(M) \\
\hline
\hline
MS-TCN \cite{mstcn} & 86.47 & 83.73 & 71.20 & 79.92 & 76.53 & 0.80 & 0.80 \\
ASRF \cite{asrf} & 86.52 & 85.27 & 73.68 & 78.59 & 76.01 & 1.30 & 1.30 \\
BCN \cite{bcn} & 88.07 & \textbf{85.88} & \textbf{75.73} & 82.26 & \textbf{79.95} & 16.64 & 16.66 \\
MS-TCN + Ours & \textbf{88.13} & 84.62  & 72.40 & \textbf{83.23} & 78.20 & \textbf{0.80} & \textbf{0.80} \\
\hline
\end{tabular}
}
\caption{
\textbf{Comparison with existing boundary-aware methods (ASRF, BCN) and the proposed method on GTEA.}
We report performance metrics along with computational cost (MACs) and model size (Params).
}
\label{table3}
\end{table*}

\subsection{Experimental Results}
\label{sub:results}

\subsubsection{Quantitative results.}
\label{subsub:quant}

Table \ref{table2} reports the performance of the proposed method applied to MS-TCN, C2F-TCN, and FACT on three benchmarks: GTEA, 50Salads, and Breakfast.
Overall, the proposed method improved Edit and F1 scores across multiple backbones and datasets, while frame-wise accuracy remained largely comparable. 
% ****** FACT on Breakfastは実験中？→現時点での最高値を追加しました。

For MS-TCN, the gains are most evident on 50Salads, with improvements of \textbf{+4.0\%} in Edit and \textbf{+3.5\%} in F1@10.  
On GTEA, the proposed method also improved segmental coherence, reflected by a \textbf{+3.3\%} increase in Edit.  
On the large-scale dataset, Breakfast, the proposed method also showed consistent improvements, confirming that the trends observed on smaller datasets generalize to more challenging large-scale scenarios.  
In particular, F1@10 achieved the largest gain across all datasets, with an improvement of \textbf{+5.4\%}.
For FACT, despite its stronger baseline, the proposed method still yielded further benefits, such as \textbf{+1.5\%} in Edit on GTEA.
However, on Breakfast, since FACT already attains a high baseline performance, the relative gains were more limited.

We also analyzed the interaction with C2F-TCN, which employs its own coarse-to-fine refinement architecture.
On GTEA, our method provides a significant boost over the strong baseline, improving Edit by \textbf{+4.6\%} and increasing F1 scores across thresholds.
On 50Salads and Breakfast, we observe consistent gains in Edit and modest improvements in most F1 metrics, while some settings show marginal drops in frame-wise accuracy (and F1@25 on 50Salads).
These results suggest that the proposed losses primarily enhance segment-level coherence, and their impact on frame-wise accuracy can be model- and dataset-dependent.

Overall, the proposed method improves Edit and F1 scores across multiple backbones and datasets, while frame-wise accuracy remains largely comparable.

\subsubsection{Qualitative results.}
\label{subsub:qual}

\begin{figure}[t]
\centering
\includegraphics[width=0.9\columnwidth]{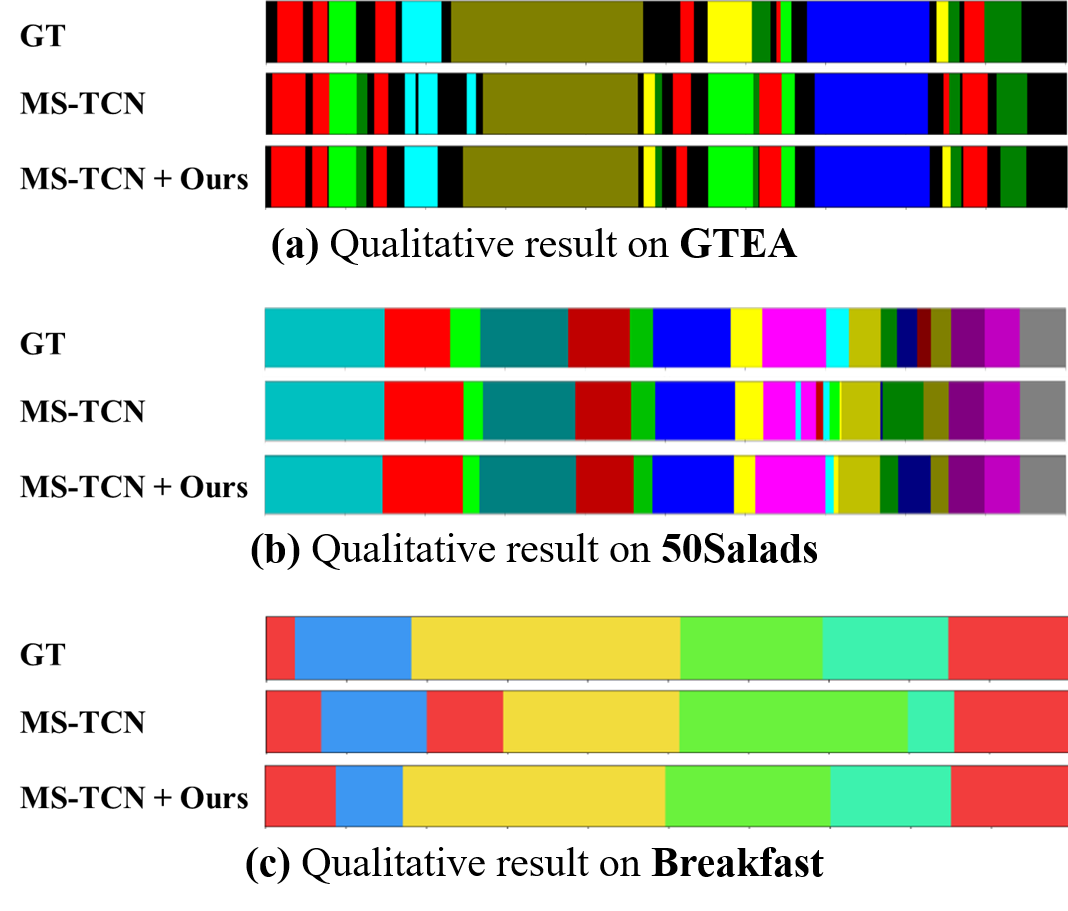} 
\caption{\textbf{Qualitative comparison on GTEA, 50Salads, and Breakfast} using MS-TCN as the backbone.
For each dataset, we show the ground-truth (GT), baseline MS-TCN predictions, and MS-TCN trained with the proposed losses.
}
\label{fig5}
\end{figure}

Figure \ref{fig5} shows qualitative comparisons between the baseline MS-TCN and the proposed method on three datasets: (a) GTEA, (b) 50Salads, and (c) Breakfast.  
In all cases, incorporating the proposed losses led to temporally smoother predictions that better aligned with the ground-truth.
In Figure \ref{fig5} (a), MS-TCN with the proposed method successfully suppressed spurious transitions and recovered the correct segment boundaries.  
The over-segmentation observed in the baseline was notably reduced.  
A clear improvement can be observed in the first half, where fragmented label transitions are effectively suppressed.

In Figure \ref{fig5} (b), our method produced more coherent long-range predictions and alleviated fragmented outputs in the latter half of the sequence.  
This demonstrates that the CDF-based segment shape regularization loss promotes segment-level consistency, while the Boundary-Regression Loss enhances transition localization.
In Figure \ref{fig5} (c), our method also improved the alignment of long-duration actions. 

Overall, these visualizations support our quantitative findings and highlight the practical advantages of the proposed training losses in improving both boundary precision and the structural integrity of predicted action sequences.

\subsubsection{Comparison with Boundary-Aware Methods.}
\label{subsub:compare_boundary}

\begin{figure}[t]
\centering
\includegraphics[width=0.9\columnwidth]{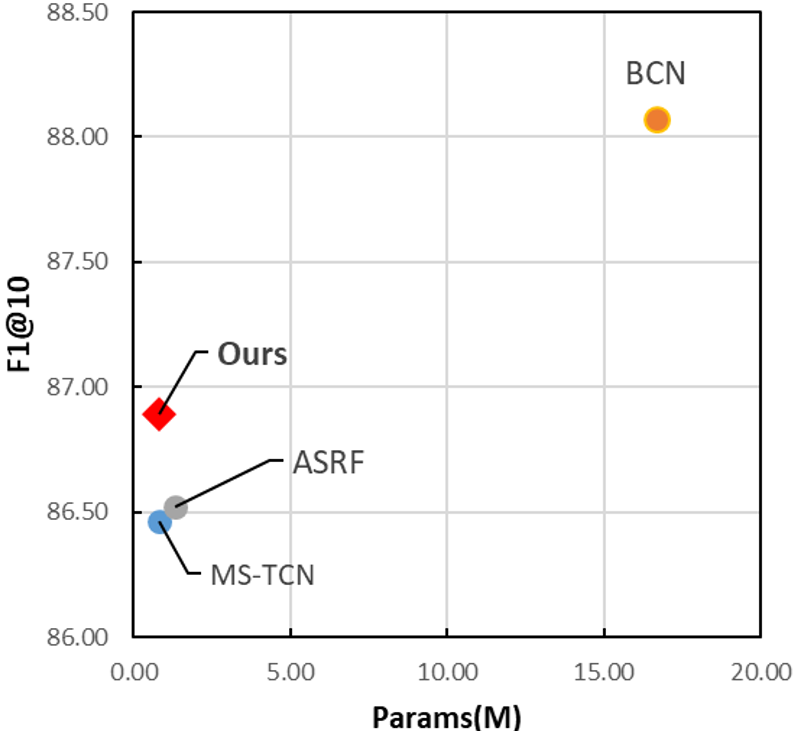} 
\caption{
The \textbf{trade-off} between Params(M) and F1@10 for conventional boundary-aware methods and our method, on GTEA.
}
\label{fig7}
\end{figure}

Table \ref{table3} and Figure \ref{fig7} present a comparison of our method with prior boundary-aware approaches, ASRF \cite{asrf} and BCN \cite{bcn}, on GTEA.
Although all methods are built on MS-TCN architecture, they differ significantly in how they incorporate boundary information and in their computational cost.

BCN achieved the highest Acc and F1 score at all thresholds, but at a substantial computational cost, over 16$\times$ MACs and parameters compared to MS-TCN.
ASRF also introduces a boundary regression branch, resulting in a moderate increase in complexity.
Although we carefully followed the official implementation, our reproduced results did not fully match the scores reported in the original paper.
Nevertheless, the observed trend is consistent: ASRF provides only limited gains over the baseline despite its added complexity.

In contrast, our method improved the performance across all metrics compared to the MS-TCN, with a particularly notable improvement in Edit (+3.31\%), indicating enhanced temporal coherence.
Importantly, this is achieved with only negligible computational and parameter overhead, since the proposed method introduces just one additional output channel as a minimal architectural modification.

Figure \ref{fig7} shows that the proposed method achieved a favorable trade-off between performance and efficiency among boundary-aware approaches.
This suggests that our loss-based approach is more efficient in leveraging boundary information than methods relying on additional modules or branches.

These results support the conclusion that the proposed losses provide an efficient way to incorporate boundary-aware supervision, offering a compelling alternative to heavyweight boundary regression networks.

\subsection{Ablation Study}
\label{sub:ablation}

\subsubsection{Component Analysis.}

Table \ref{table4} presents an ablation study on the GTEA dataset using MS-TCN, designed to evaluate the individual and joint effects of the proposed loss functions.
Adding the Boundary-Regression Loss ($\mathcal{L}_{\text{B}}$) alone improved the performance across all metrics, especially in Edit distance and Acc, suggesting enhanced boundary localization and temporal coherence.
On the other hand, adding only Segment Shape Regularization Loss ($\mathcal{L}_{\text{S}}$) yielded the highest F1@25 and F1@50, indicating that it effectively improves intra-segment consistency and corrects local misalignments.

When both losses are adopted together using our temporally decoupled loss assignment, the model achieved the best overall performance, including the highest Edit distance and Acc.
These results validated that the two losses address complementary aspects of temporal action segmentation—boundary sharpness and segment-level structure—and their combination leads to more accurate and temporally consistent predictions.

\begin{table}[t]
\centering
\resizebox{0.9\columnwidth}{!}{
\begin{tabular}{ccccccc}
\hline
$\mathcal{L}_{\text{B}}$ & $\mathcal{L}_{\text{S}}$ & \multicolumn{3}{c}{F1@\{10,25,50\}} & Edit & Acc \\
\hline
\hline
  & & 86.47 & 83.73 & 71.20 & 79.92 & 76.53  \\
 $\surd$ & & 87.84 & 85.24 & 72.16 & 82.83 & 77.69 \\
  & $\surd$ & 87.36 & \textbf{85.55} & \textbf{72.73} & 82.42 & 77.30 \\
 $\surd$ & $\surd$ & \textbf{88.13} & 84.62  & 72.40 & \textbf{83.23} & \textbf{78.20} \\
\hline
\end{tabular}
}
\caption{\textbf{Ablation study on GTEA with MS-TCN.}
We evaluate the individual and joint contributions of $\mathcal{L}_{\text{B}}$ and $\mathcal{L}_{\text{S}}$. 
}
\label{table4}
\end{table}

\subsubsection{Start Epoch Sensitivity.}

Table \ref{table5} investigates the sensitivity of Segment Shape Regularization Loss to its start epoch $E_{\text{start}}$ during training on the GTEA dataset with MS-TCN.
The total number of training epochs is 50, and all other components and training conditions were kept fixed.

We observe that applying $\mathcal{L}_{\text{S}}$ after epoch 20 yields the best overall performance in terms of F1@10, F1@50, and Acc.
Although starting earlier ($E_{\text{start}}=10$) results in a slightly higher Edit distance, it comes at the cost of reduced F1 score and Acc, likely due to unstable predictions in the early training phase interfering with segment-level alignment.
Conversely, delaying $\mathcal{L}_{\text{S}}$ to epoch 30 slightly improves F1@25, but degrades Edit distance and F1@50, suggesting missed opportunities for refining segment structure during mid-stage learning.

These results validated our hypothesis that $\mathcal{L}_{\text{S}}$ is most effective when applied after the model has achieved coarse classification, and highlight the importance of scheduling segment-level regularization during training.

We confirmed a similar trend on 50Salads and Breakfast, where introducing $\mathcal{L}_{\text{S}}$ too early similarly degraded performance due to unstable early predictions.
Therefore, for all datasets with MS-TCN, we fixed $E_{\text{start}}=20$ as a consistent and empirically optimal setting.
For C2F-TCN and FACT, $E_{\text{start}}$ was also tuned via cross-validation to account for their different convergence behaviors.

\begin{table}[t]
\centering
\resizebox{0.9\columnwidth}{!}{
\begin{tabular}{cccccc}
\hline
$E_{\text{start}}$ & \multicolumn{3}{c}{F1@\{10,25,50\}} & Edit & Acc \\
\hline
\hline
  0 & 81.53 & 77.35 & 60.63 & 75.84 & 75.13 \\
 10 & 86.45 & 84.43 & 72.37 & \textbf{83.62} & 76.16 \\
 20 & \textbf{88.13} & 84.62  & \textbf{72.40} & 83.23 & \textbf{78.20} \\
 30 & 86.86 & \textbf{84.81}  & 71.71 & 81.71 & 77.77 \\
\hline
\end{tabular}
}
\caption{
\textbf{Ablation study on the start epoch $E_{\text{start}}$ for applying Segment Shape Regularization Loss on GTEA using MS-TCN.  }
We evaluate how the timing of introducing $\mathcal{L}_{\text{S}}$ affects performance, while keeping all other settings identical.
}
\label{table5}
\end{table}

\subsubsection{Effect of Temporally Decoupled Loss Assignment.}

Table \ref{table6} compares two loss assignment strategies:
(1) "All frames" applies both $\mathcal{L}_{\text{B}}$ and $\mathcal{L}_{\text{S}}$ to all frames uniformly, and
(2) "Ours" applies them in a temporally decoupled manner based on boundary proximity.

We observe that our proposed boundary-aware strategy yields higher Edit distance and Acc while the uniform setting achieves competitive F1 scores.
This suggests that separating the two losses reduces gradient conflict and better balances sharp boundary localization with smooth segment-level alignment.
The improvement in structural metrics (Edit and Acc) confirms the benefit of temporally targeted regularization over naive joint supervision.

\begin{figure}[t]
\centering
\includegraphics[width=1.\columnwidth]{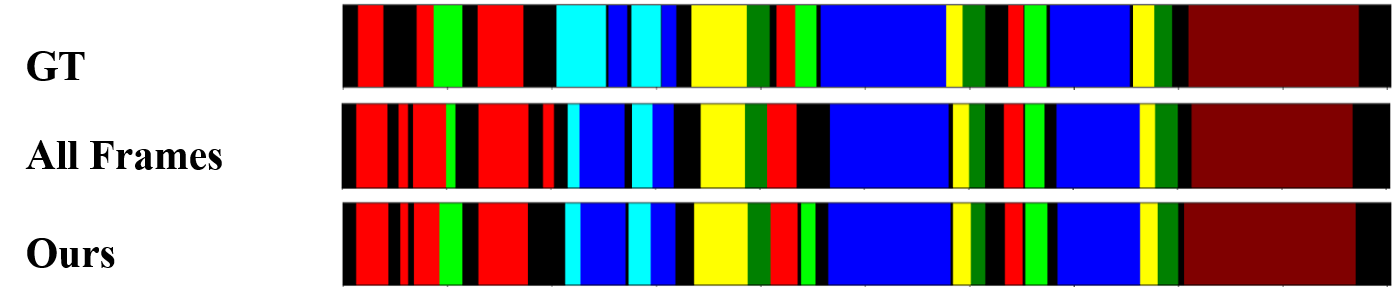} 
\caption{\textbf{Qualitative results of “All frames” vs. “Ours” on GTEA.}
Boundary-aware loss assignment improves alignment and segmentation quality, especially near action boundaries.
}
\label{fig8}
\end{figure}

Figure \ref{fig8} further supports this observation, showing qualitative predictions on a GTEA sequence.
Compared to the "All frames" model, our method produces more coherent and semantically meaningful action segments, especially around complex transitions, by mitigating over-segmentation and reducing spurious boundaries.

\begin{table}[t]
\centering
\resizebox{0.9\columnwidth}{!}{
\begin{tabular}{lccccc}
\hline
 Methods & \multicolumn{3}{c}{F1@\{10,25,50\}} & Edit & Acc \\
\hline
\hline
 All frames & 87.49 & \textbf{85.08} & \textbf{72.44} & 81.61 & 77.64 \\
 Ours & \textbf{88.13} & 84.62  & 72.40 & \textbf{83.23} & \textbf{78.20} \\
\hline
\end{tabular}
}
\caption{
\textbf{Ablation study on loss assignment strategy using MS-TCN on GTEA}.  
"All frames" applies both $\mathcal{L}_{\text{B}}$ and $\mathcal{L}_{\text{S}}$ to all frames uniformly, while "Ours" follows our proposed boundary-aware strategy:  
$\mathcal{L}_{\text{B}}$ is applied only near ground-truth boundaries, and $\mathcal{L}_{\text{S}}$ is applied to non-boundary regions.
}
\label{table6}
\end{table}

\begin{table*}[t]
\centering
\resizebox{2.\columnwidth}{!}{
\begin{tabular}{lccccc | ccccc | ccccc}
\hline
Dataset & \multicolumn{5}{c|}{GTEA} & \multicolumn{5}{c}{50Salads} & \multicolumn{5}{c}{Breakfast} \\
\hline
Methods & \multicolumn{3}{c}{F1@\{10,25,50\}} & Edit & Acc & \multicolumn{3}{c}{F1@\{10,25,50\}} & Edit & Acc & \multicolumn{3}{c}{F1@\{10,25,50\}} & Edit & Acc \\
\hline
\hline
MS-TCN \cite{mstcn} 
& 86.47 & 83.73 & 71.20 & 79.92 & 76.53 
& 73.96 & 71.42 & 62.12 & 66.29 & 79.11
& 50.61 & 46.41 & 36.58 & 61.75 & 66.96 \\
+ Ours 
& 88.13 & 84.62  & 72.40 & 83.23 & 78.20 
& 77.44 & 75.22 & 67.31 & 70.29 & 81.17 
& \textbf{55.96} & \textbf{51.39} & \textbf{40.78} & \textbf{62.29} & \textbf{67.46} \\
+ ASOT \cite{asot} 
& 87.81 & 85.49 & 73.30 & 81.65 & 76.61
& 81.36 & 79.17 & 70.16 & 74.30 & 79.49
& 46.91 & 42.50 & 32.01 & 44.94 & 61.10 \\
+ ASOT + $\mathcal{L}_{\text{B}}$ 
& 88.12 & 85.99 & 73.95 & 83.33 & 78.04
& \textbf{82.78} & 80.97 & 71.69 & 74.77 & 81.27
& 45.13 & 40.89 & 30.36 & 42.89 & 59.41 \\
+ ASOT + $\mathcal{L}_{\text{S}}$ 
& 88.03 & 84.92 & 73.44 & 82.01 & 77.03
& 79.78 & 77.79 & 69.65 & 71.86 & 78.43
& 49.25 & 44.80 & 34.28 & 46.33 & 61.90 \\
+ ASOT + Ours 
& \textbf{88.82} & \textbf{87.09} & \textbf{74.31} & \textbf{84.88} & \textbf{78.48}
& 82.45 & \textbf{81.05} & \textbf{73.00} & \textbf{76.08} & \textbf{81.48}
& 45.42 & 41.18 & 31.24 & 44.71 & 61.23 \\
\hline
\end{tabular}
}
\caption{\textbf{Ablation study on synergy with post-processing method} using MS-TCN on GTEA, 50Salads and Breakfast. 
We apply ASOT \cite{asot} as a post-processing step to both the baseline and our model trained with the proposed losses.
}
\label{table7}
\end{table*}

\subsubsection{Synergy with Post-Processing Methods}

Beyond analyzing the internal components of our method, we also investigate its synergy with external post-processing methods. 
In particular, we focus on ASOT \cite{asot}, which was originally proposed for unsupervised TAS but can be applied as a post-processing refinement in supervised settings. 
We applied ASOT to the MS-TCN and to models trained with our framework and its individual components.

Table \ref{table7} demonstrated a clear synergistic relationship on GTEA and 50Salads.
On GTEA, the effect is evident. 
Although both the proposed method alone ("+ Ours") and ASOT alone ("+ ASOT") improved upon the baseline, their combination ("+ ASOT + Ours") achieved the best performance across all metrics. 
Notably, the combined model reaches an F1@10 of 88.82 and an Edit of 84.88, surpassing both the proposed method alone and ASOT alone.

A similar and even more pronounced trend is observed on 50Salads. 
Both our method alone and ASOT alone provide significant gains over the baseline. 
However, combining them ("+ ASOT + Ours") yields the strongest overall results.
This combination achieved an F1@10 of 82.45 and an F1@50 of 73.00, which are substantial improvements over both the proposed method alone and ASOT alone.

In contrast, a different trend was observed on Breakfast.
As shown in Table \ref{table7}, applying ASOT alone ("+ ASOT") significantly degraded performance compared to the baseline, and even combining ASOT with either $\mathcal{L}_{\text{B}}$, $\mathcal{L}_{\text{S}}$, or our full method failed to recover this drop.
This indicates that ASOT is inherently ineffective on Breakfast, likely because its post-hoc temporal refinement conflicts with the long, diverse action sequences in this dataset.
In contrast, our method ("+ Ours") improved the baseline on all metrics, while ASOT-based post-processing is not beneficial.

These ablation studies collectively demonstrate that the proposed method not only improves predictions through its design, including complementary losses and targeted supervision, but also provides a stronger foundation for external refinements, such as ASOT. 
This highlights the versatility of the proposed method as both an internal regularizer and a complementary training scheme for post-hoc methods.
\section{Conclusion}
\label{sec:conclusion}

We presented a simple training-time enhancement for fully supervised Temporal Action Segmentation (TAS) based on two complementary auxiliary losses: a Boundary-Regression Loss and a CDF-based Segment Shape Regularization Loss.
Unlike prior boundary-aware approaches that rely on additional branches or complex architectural modifications, our method is architecture-agnostic and requires only minimal changes: one additional boundary output channel, and two loss terms added to the original loss.

Experiments on GTEA, 50Salads, and Breakfast demonstrated that the proposed losses improved Edit and F1 scores across MS-TCN, C2F-TCN, and FACT in many settings, while frame-wise accuracy remained largely comparable.
We further show that the two losses address complementary aspects of TAS, boundary localization, and within-segment structural consistency, and that applying them with a temporally decoupled loss assignment helps mitigate gradient conflicts.
In addition, compared to boundary-aware baselines such as ASRF and BCN, our approach achieves a favorable accuracy--efficiency trade-off with negligible computational overhead.
Finally, combining our losses with ASOT as a post-hoc refinement can provide additional gains on certain datasets, suggesting that training-time regularization and inference-time refinement can be complementary.

Overall, our results indicated that improving boundary fidelity and segment-level structure through loss design is a practical direction for enhancing TAS.
We hope this work encourages further exploration of lightweight, architecture-agnostic objectives for temporal segmentation and sequence labeling tasks.
{
    \small
    \bibliographystyle{ieeenat_fullname}
    \bibliography{main}
}


\end{document}